\documentclass[11pt,a4paper]{article}
\usepackage[hyperref]{acl2020}
\usepackage{times}
\usepackage{latexsym}

\usepackage{microtype}

\aclfinalcopy 
\def\aclpaperid{2987} 

\usepackage{bm, color, multirow, graphicx, amsmath, amssymb}

\newcommand\eg{\textit{e.g.}}
\newcommand\ie{\textit{i.e.}}
\newcommand\etc{\textit{etc}}
\newcommand\wrt{\textit{w.r.t}}

\title{Single-/Multi-Source Cross-Lingual NER via Teacher-Student Learning on Unlabeled Data in Target Language}

\author{Qianhui Wu$^1$, Zijia Lin$^2$, B\"orje F. Karlsson$^2$, Jian-Guang Lou$^2$, and Biqing Huang$^1$ \\
	$^1$Beijing National Research Center for Information Science and Technology (BNRist) \\
	Department of Automation, Tsinghua University, Beijing 100084, China \\
	\texttt{wuqianhui@tsinghua.org.cn}, \texttt{hbq@tsinghua.edu.cn} \\
	$^2$Microsoft Research, Beijing 100080, China \\
	\texttt{\{zijlin,borje.karlsson,jlou\}@microsoft.com} 
	}
\date{}

\begin{document}
	\maketitle
	\begin{abstract}
		To better tackle the named entity recognition (NER) problem on languages with little/no labeled data, cross-lingual NER must effectively leverage knowledge learned from source languages with rich labeled data. 
		Previous works on cross-lingual NER are mostly based on label projection with pairwise texts or direct model transfer. 
		However, such methods either are not applicable if the labeled data in the source languages is unavailable, or do not leverage information contained in unlabeled data in the target language.
		In this paper, we propose a teacher-student learning method to address such limitations, where NER models in the source languages are used as teachers to train a student model on unlabeled data in the target language. The proposed method works for both single-source and multi-source cross-lingual NER. 
		For the latter, we further propose a similarity measuring method to better weight the supervision from different teacher models. 
		Extensive experiments for 3 target languages on benchmark datasets well demonstrate that our method outperforms existing state-of-the-art methods for both single-source and multi-source cross-lingual NER.
	\end{abstract}

	\section{Introduction}
	Named entity recognition (NER) is the task of identifying text spans that belong to pre-defined categories, like locations, person names, \etc. 
	It's a fundamental component in many downstream tasks, and has been greatly advanced by deep neural networks~\cite{lample2016neural,chiu2016named,peters2017semi}.
	However, these approaches generally require massive manually labeled data, which prohibits their adaptation to low-resource languages due to high annotation costs. 
	
	One solution to tackle that is to transfer knowledge from a source language with rich labeled data to a target language with little or even no labeled data, which is referred to as cross-lingual NER~\cite{wu2019beto,wu2020enhanced}.
	In this paper, following~\citet{wu2019beto} and \citet{wu2020enhanced}, we focus on the extreme scenario of cross-lingual NER where \textbf{no labeled data} is available in the target language, which is challenging in itself and has attracted considerable attention from the research community in recent years.
	
	\begin{figure}[t]
		\centering
		\includegraphics[width=7cm]{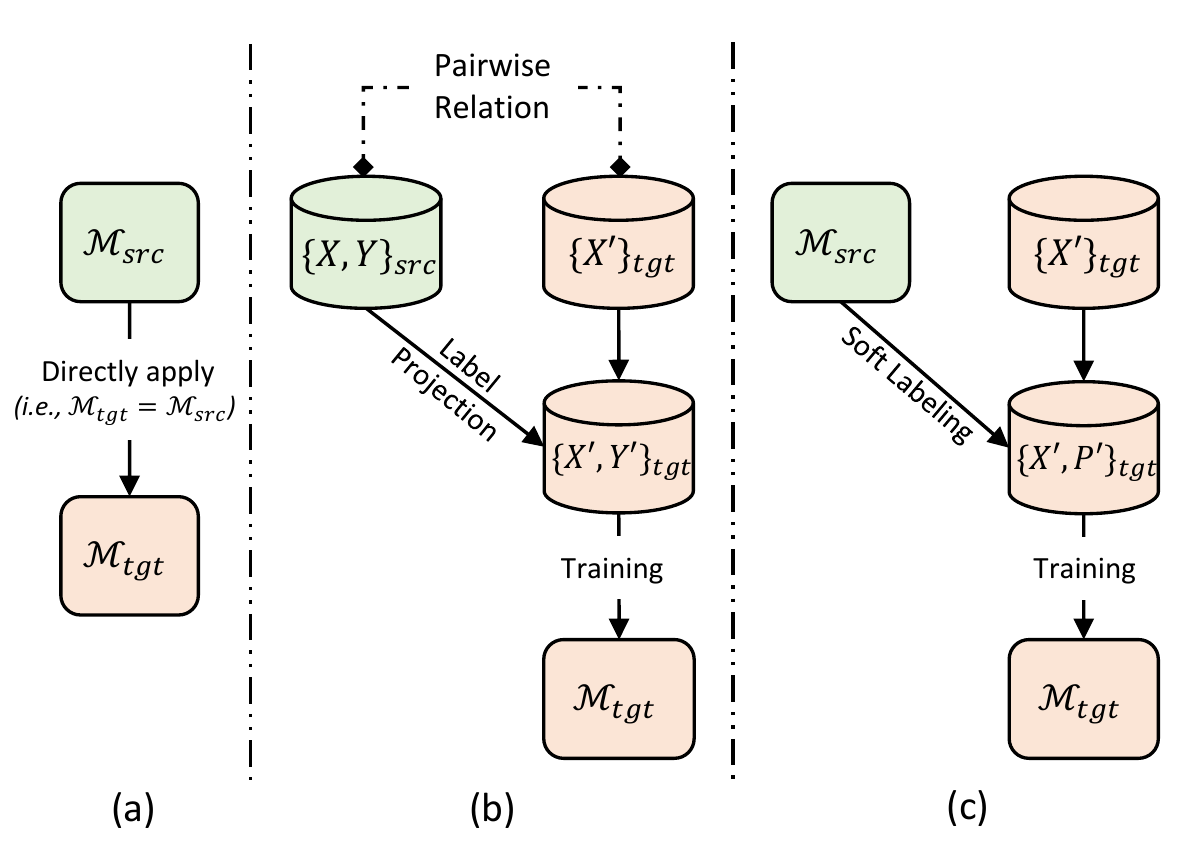}
		\caption{Comparison between previous cross-lingual NER methods (\textbf{a}/\textbf{b}) and the proposed method (\textbf{c}). \textbf{(a)}: direct model transfer; \textbf{(b)}: label projection with pairwise texts; \textbf{(c)}: proposed teacher-student learning method. 
			$\mathcal{M}_{src/tgt}$: learned NER model for source/target language; $\{X,Y\}_{src}$: labeled data in source language; $\{X'\}_{tgt}$: unlabeled data in target language; $\{X', Y'\}_{tgt}/\{X', P'\}_{tgt}$: pseudo-labeled data in target language with hard labels / soft labels. 
		}
		\label{fig:motivation}
	\end{figure}
	
	Previous works on cross-lingual NER are mostly based on label projection with pairwise texts or direct model transfer. Label-projection based methods focus on using labeled data in a source language to generate pseudo-labelled data in the target language for training an NER model. 
	For example, \citet{ni2017weakly} creates automatically labeled NER data for the target language via label projection on comparable corpora and develops a heuristic scheme to select good-quality projection-labeled data.
	\citet{mayhew2017cheap} and \citet{xie2018neural} translate the source language labeled data at the phrase/word level to generate pairwise labeled data for the target language. 
	Differently, model-transfer based methods~\cite{wu2019beto, wu2020enhanced} focus on training a shared NER model on the labeled data in the source language with language-independent features, such as cross-lingual word representations~\cite{devlin2019bert}, and then directly testing the model on the target language.
	
	However, there are limitations in both label-projection based methods and model-transfer based methods. 
	The former relies on labeled data in the source language for label projection, and thus is not applicable in cases where the required labeled data is inaccessible (\eg, due to privacy/sensitivity issues).  
	Meanwhile, the later does not leverage unlabeled data in the target language, which can be much cheaper to obtain and probably contains very useful language information. 
	
	In this paper, we 
	propose a teacher-student learning method for cross-lingual NER to address the mentioned limitations. 
	Specifically, we leverage multilingual BERT~\cite{devlin2019bert} as the base model to produce language-independent features. A previously trained NER model for the source language is then used as a teacher model to predict the probability distribution of entity labels (\ie, soft labels) for each token in the \textit{non-pairwise} unlabeled data in the target language. Finally, we train a student NER model for the target language using the pseudo-labeled data with such soft labels.   
	The proposed method does not rely on labelled data in the source language, and it also leverages the available information from unlabeled data in the target language, thus avoiding the mentioned limitations of previous works. 
	Note that we use the teacher model to predict soft labels rather than hard labels (\ie, one-hot labelling vector), as soft labels can provide much more information~\cite{hinton2015distilling} for the student model. 
	Figure~\ref{fig:motivation} shows the differences between the proposed teacher-student learning method and the typical label-projection or model-transfer based methods. 
	
	We further extend our teacher-student learning method to multi-source cross-lingual NER, considering that there are usually multiple source languages available in practice and we would prefer transferring knowledge from all source languages rather than a single one. 
	In this case, our method still enjoys the same advantages in terms of data availability and inference efficiency, compared with existing works~\cite{tackstrom2012nudging,chen2019multi,enghoff2018low, rahimi2019massively}. 
	Moreover, 
	we propose a method to measure the similarity between each source language and the target language, and use this similarity to better weight the supervision from the corresponding teacher model. 
	
	We evaluate our proposed method for 3 target languages on benchmark datasets, using different source language settings. Experimental results show that our method outperforms existing state-of-the-art methods for both single-source and multi-source cross-lingual NER. We also conduct case studies and statistical analyses to discuss why teacher-student learning reaches better results.
	
	The main contributions of this work are:
	\begin{itemize}
		\item We propose a teacher-student learning method for single-source cross-lingual NER, which addresses limitations of previous works \wrt{} data availability and usage of unlabeled data.
		\item We extend the proposed method to multi-source cross-lingual NER, using a measure of the similarities between source/target languages to better weight teacher models.
		\item We conduct extensive experiments validating the effectiveness and reasonableness of the proposed methods, and further analyse why they attain superior performance. 
	\end{itemize}
	
	\section{Related Work}
	\paragraph{Single-Source Cross-Lingual NER: }
	Such approaches consider one single source language for knowledge transfer. Previous works can be divided into two categories: label-projection and model-transfer based methods.
	
	Label-projection based methods aim to build pseudo-labeled data for the target language to train an NER model. 
	Some early works proposed to use bilingual parallel corpora and project model expectations~\cite{wang2014cross} or labels~\cite{ni2017weakly} from the source language to the target language with external word alignment information. 
	But obtaining parallel corpora is expensive or even infeasible.
	To tackle that, recent methods proposed to firstly translate source-language labeled data at the phrase level~\cite{mayhew2017cheap} or word level~\cite{xie2018neural}, and then directly copy labels across languages. But translation introduces extra noise due to sense ambiguity and word order differences between languages, thus hurting the trained model.
	
	Model-transfer based methods generally rely on language-independent features (\eg, cross-lingual word embeddings~\cite{ni2017weakly,huang2019cross,wu2019beto,moon2019lingua}, word clusters~\cite{tackstrom2012cross}, gazetteers~\cite{zirikly2015cross}, and \emph{wikifier} features~\cite{tsai2016cross}), so that a model trained with such features can be directly applied to the target language. 
	For further improvement, \citet{wu2020enhanced} proposed constructing a pseudo-training set for each test case and fine-tuning the model before inference. However, these methods do not leverage any unlabeled data in the target language, though such data can be easy to obtain and benefit the language/domain adaptation. 
	
	\paragraph{Multi-Source Cross-Lingual NER: }
	Multi-source cross-lingual NER considers multiple source languages for knowledge transfer.
	
	\citet{tackstrom2012nudging} and \citet{moon2019lingua} concatenated the labeled data of all source languages to train a unified model, and performed cross-lingual NER in a direct model transfer manner. \citet{chen2019multi} leveraged adversarial networks to learn language-independent features, and learns a mixture-of-experts model~\cite{shazeer2017outrageously} to weight source models at the token level. However, both methods straightly rely on the availability of labeled data in the source languages.
	
	Differently, \citet{enghoff2018low} implemented multi-source label projection and studied how source data quality influence performance. 
	\citet{rahimi2019massively} applied truth inference to model the transfer annotation bias from multiple source-language models. However, both methods make predictions via an ensemble of source-language models, which is cumbersome and computationally expensive, 
	especially when a source-language model has massive parameter space.
	
	\paragraph{Teacher-Student Learning: }
	Early applications of teacher-student learning targeted model compression~\cite{bucilu2006model}, where a small student model is trained to mimic a pre-trained, larger teacher model or ensemble of models. It was soon applied to various tasks like image classification~\cite{hinton2015distilling,you2017learning}, dialogue generation~\cite{peng2019teacher}, and neural machine translation~\cite{tan2019multilingual}, which demonstrated the usefulness of the knowledge transfer approach.
	\newline
	\newline
	In this paper, we investigate teacher-student learning for the task of cross-lingual NER, in both single-source and multi-source scenarios. Different from previous works, our proposed method does not rely on the availability of labelled data in source languages or any pairwise texts, while it can also leverage extra information in unlabeled data in the target language to enhance the cross-lingual transfer. Moreover, compared with using an ensemble of source-language models, our method uses a single student model for inference, which can enjoy higher efficiency.
	
	
	\section{Methodology}
	Named entity recognition can be formulated as a sequence labeling problem, \ie, given a sentence $\bm{x} = \{ x_i \}_{i=1}^L$ with $L$ tokens, an NER model is supposed to infer the entity label $y_i$ for each token $x_i$ and output a label sequence $\bm{y} = \{ y_i \}_{i=1}^L$. 
	Under the paradigm of cross-lingual NER, we assume there are $K$ source-language models previously trained with language-independent features. 
	Our proposed teacher-student learning method then uses those $K$ source-language models as teachers to train an effective student NER model for the target language on its unlabeled data $D_{tgt}$. 
	
	\subsection{Single-Source Cross-Lingual NER}
\begin{figure}[t]
		\centering
		\includegraphics[width=7cm]{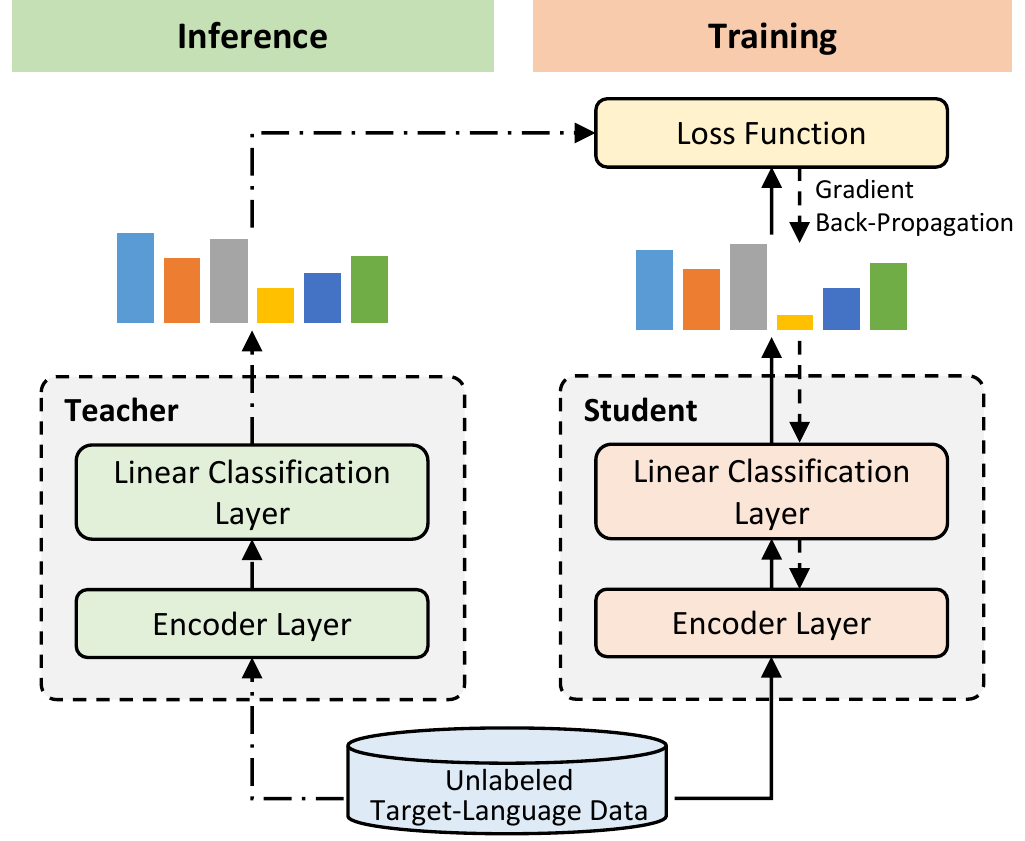}
		\caption{Framework of the proposed teacher-student learning method for \textbf{single-source} cross-lingual NER.}
		\label{fig:frame_single_source}
	\end{figure}

	Here we firstly consider the case of only one source language ($K=1$) for cross-lingual NER. 
	The overall framework of the proposed teacher-student learning method for single-source cross-lingual NER is illustrated in Figure~\ref{fig:frame_single_source}. 
	
	\subsubsection{NER Model Structure}
	\label{sec:ner_model}
	As shown in Figure~\ref{fig:frame_single_source}, for simplicity, we employ the same neural network structure for both teacher (source-language) and student (target-language) NER models.
	Note that the student model is flexible and its structure can be determined according to the trade-off between performance and training/inference efficiency. 
	
	Here the adopted NER model consists of an encoder layer and a linear classification layer. Specifically, given an input sequence $\bm{x}=\{x_i\}_{i=1}^L$ with $L$ tokens, the encoder layer $f_{\theta}$ maps it into a sequence of hidden vectors $\bm{h}=\{h_i\}_{i=1}^L$:
	\begin{equation}
	\label{equ:h_single}
	\bm{h} = f_{\theta}(\bm{x})
	\end{equation}
	Here $f_{\theta}(\cdot)$ can be any encoder model that produces cross-lingual token representations, and $h_i$ is the hidden vector corresponding to the $i$-th token $x_i$.
	
	With each $h_i$ derived, the linear classification layer computes the probability distribution of entity labels for the corresponding token $x_i$, using a \textit{softmax} function:
	\begin{equation}
	\label{equ:p_single}
	p(x_i, \Theta) = \text{softmax}(Wh_i+b)
	\end{equation}
	where $p(x_i, \Theta)\in \mathbb{R}^{|C|}$ with $C$ being the entity label set, and $\Theta=\{f_{\theta}, W, b\}$ denotes the to-be-learned model parameters.
	
	\subsubsection{Teacher-Student Learning}
	\paragraph{Training: } We train the student model to mimic the output probability distribution of entity labels by the teacher model, on the unlabeled data in the target language $D_{tgt}$. 
	Knowledge from the teacher model is expected to transfer to the student model, while the student model can also leverage helpful language-specific information available in the unlabeled target-language data.
	
	Given an unlabeled sentence $\bm{x'} \in D_{tgt}$ in the target language, the teacher-student learning loss \wrt{}  $\bm{x'}$ is formulated as the \textit{mean squared error} (MSE) between the output probability distributions of entity labels by the student model and those by the teacher model, averaged over tokens. 
	Note that here we follow~\citet{yang2019model} and use the MSE loss, because it is symmetric and mimics all probabilities equally.
	Suppose that for the $i$-token in $\bm{x'}$, \ie, $x'_i$, the probability distribution of entity labels output by the student model is denoted as $\hat{p}(x'_i, \Theta_S)$, and that output by the teacher model as $\tilde{p}(x'_i, \Theta_T)$. Here $\Theta_S$ and $\Theta_T$, respectively, denote the  parameters of the student and the teacher models. 
	The teacher-student learning loss \wrt{} $\bm{x'}$ is then defined as:
	\begin{equation}
	\label{equ:loss}
	\mathcal{L}(\bm{x'}, \Theta_S) = \frac{1}{L}\sum_{i=1}^L\text{MSE}\left( \hat{p}(x'_i, \Theta_S), \tilde{p}(x'_i, \Theta_T) \right)
	\end{equation}
	And the whole training loss is the summation of losses \wrt{} all sentences in $D_{tgt}$, as defined below.
	\begin{equation}
	\label{equ:train_loss}
	\mathcal{L}(\Theta_S) = \sum_{\bm{x'} \in D_{tgt}} \mathcal{L}(\bm{x'}, \Theta_S)
	\end{equation}
	Minimizing $\mathcal{L}(\Theta_S)$ will derive the student model.
	
	\paragraph{Inference: } For inference in the target language, we only utilize the learned student model to predict the probability distribution of entity labels for each token $x_i$ in a test sentence $\bm{x}$. Then we take the entity label $c \in C$ with the highest probability as the predicted label $y_i$ for $x_i$:
	\begin{equation}
	\label{equ:infer}
	y_i = \arg\max_c \hat{p}(x_i, \Theta_S)_c
	\end{equation}
	where $p(x_i, \Theta_S)_c$ denotes the predicted probability corresponding to the entity label $c$ in $p(x_i, \Theta_S)$.

	\subsection{Multi-Source Cross-Lingual NER}
\begin{figure}[t]
		\centering
		\includegraphics[width=7cm]{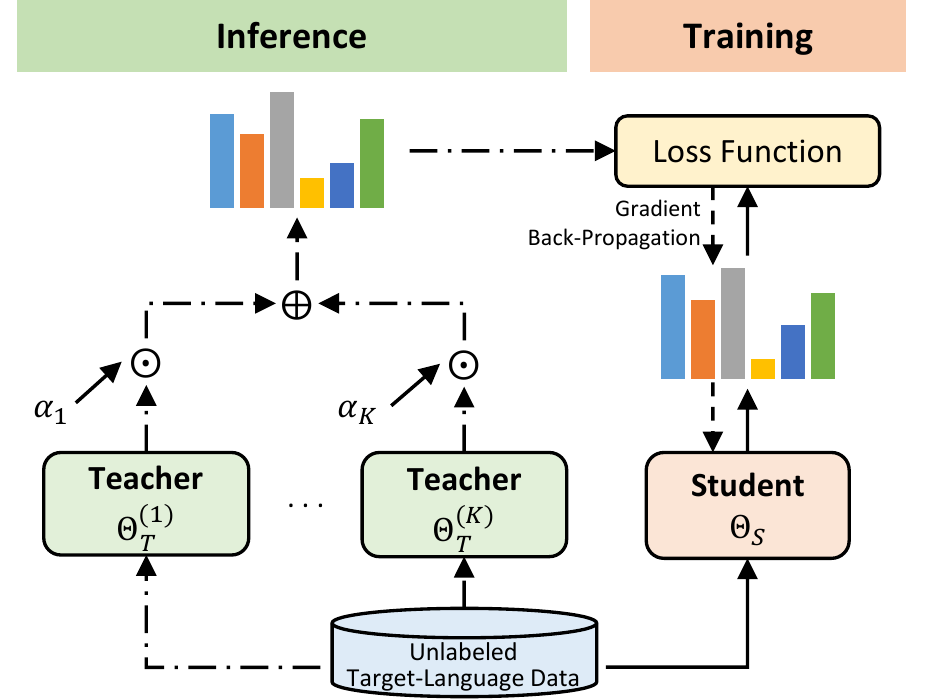}
		\caption{Framework of the proposed teacher-student learning method for \textbf{multi-source} cross-lingual NER.}
		\label{fig:frame_multi_source}
	\end{figure}

	The framework of the proposed teacher-student learning method for multi-source ($K > 1$) cross-lingual NER is illustrated in Figure~\ref{fig:frame_multi_source}. 
	
	\subsubsection{Extension to Multiple Teacher Models}
	As illustrated in Figure~\ref{fig:frame_multi_source}, 
	we extend the single-teacher framework in Figure~\ref{fig:frame_single_source} into a multi-teacher one, while keeping the student model unchanged. 
	
	Note that, for simplicity, all teacher models and the student model use the same model structure as \ref{sec:ner_model}. 
	Take the $k$-th teacher model for example, and denote its parameters as $\Theta^{(k)}_T$. 
	Given a sentence $\bm{x'}=\{x'_i\}_{i=1}^L$ with $L$ tokens from the unlabeled data $D_{tgt}$ in the target language, the output probability distribution of entity labels \wrt{} the $i$-th token $x_i$ can be derived as Eq.~\ref{equ:h_single} and \ref{equ:p_single}, which is denoted as $\tilde{p}(x'_i, \Theta^{(k)}_T)$.
	To combine all teacher models, we add up their output probability distributions with a group of weights $\{\alpha_k\}_{k=1}^K$ as follows. 
	\begin{equation}
	\label{equ:multi_teacher}
	\tilde{p}(x'_i, \Theta_T) = \sum_{k=1}^K \alpha_k \cdot \tilde{p}(x'_i, \Theta^{(k)}_T)
	\end{equation}
	where $\tilde{p}(x'_i, \Theta_T)$ is the combined probability distribution of entity labels, $\Theta_T=\{\Theta_T^{(k)}\}_{k=1}^K$ is the set of parameters of all teacher models, and $\alpha_k$ is the weight corresponding to the $k$-th teacher model, with $\sum_{k=1}^K \alpha_k = 1$ and $\alpha_k \geq 0, \forall k \in \{1, \ldots, K\}$.
	
	\subsubsection{Weighting Teacher Models}
	Here we elaborate on how to derive the weights $\{\alpha_k\}_{k=1}^K$ in cases \textit{w/} or \textit{w/o} \textbf{\textit{unlabeled}} data in the source languages. 
	Source languages more similar to the target language should generally be assigned higher weights to transfer more knowledge. 
	
	\paragraph{Without Any Source-Language Data: } 
	It is straightforward to average over all teacher models:
	\begin{equation}
	\label{equ:minimal_resource}
	\alpha_k = \frac{1}{K}, ~\forall k \in \{1, 2, \ldots, K\}
	\end{equation}
	
	\paragraph{With Unlabeled Source-Language Data: } 
	As no labeled data is available, existing supervised language/domain similarity learning methods for a target task (\ie, NER)~\cite{mcclosky2010} are not applicable here. Inspired by ~\citet{pinheiro2018unsupervised}, we propose to introduce a language identification auxiliary task for calculating similarities between source and target languages, and then weight teacher models based on this metric. 
	
	In the language identification task, for the $k$-th source language, each unlabeled sentence $\bm{u}^{(k)}$ in it is associated with the language index $k$ to build its training dataset, denoted as  $D^{(k)}_{src}=\{(\bm{u}^{(k)}, k)\}$. 
	We also assume that in the $m$-dimensional language-independent feature space, sentences from each source language should be clustered around the corresponding language embedding vector.  
	We thus introduce a learnable language embedding vector $\mu^{(k)}\in \mathbb{R}^m$ for the $k$-th source language, and then utilize a \textit{bilinear} operator to measure similarity between a given sentence $\bm{u}$ and the $k$-th source language:
	\begin{equation}
	\label{equ:similarity}
	s(\bm{u}, \mu^{(k)}) = g^T(\bm{u})M\mu^{(k)}
	\end{equation}
	where $g(\cdot)$ can be any language-independent model that outputs sentence embeddings, and $M\in \mathbb{R}^{m\times m}$ denotes the parameters of the \textit{bilinear} operator. 
	
	By building a language embedding matrix $P \in \mathbb{R}^{m\times K}$ with each $\mu^{(k)}$ \textit{column by column}, and applying a \textit{softmax} function over the \textit{bilinear} operator, we can derive language-specific probability distributions \wrt{} $\bm{u}$ as below.
	\begin{equation}
	q(\bm{u}, M,P) = \text{softmax}\left(g^T(\bm{u}) M P\right)
	\end{equation}
	Then the parameters $M$ and $P$ are trained to identify the language of each sentence in $\{D^{(k)}_{src}\}_{k=1}^K$, via minimizing the \textit{cross-entropy} (CE) loss:
	\begin{equation}
	\label{equ:lang_sim}
	\begin{aligned}
	\mathcal{L}(P, M) = &-\frac{1}{Z}\sum_{(\bm{u}^{(k)}, k) \in D_{src}} \text{CE}\left(q(\bm{u}^{(k)},M,P), k\right) \\
	&+ \gamma \Vert PP^T - I \Vert^2_F
	\end{aligned}
	\end{equation}
	where $D_{src}$ is the union set of $\{D^{(k)}_{src}\}_{k=1}^K$, $Z=|D_{src}|$, $\Vert \cdot \Vert_F^2$ denotes the squared Frobenius norm, and $I$ is an identity matrix. The regularizer in $\mathcal{L}(P, M)$ is to encourage different dimensions of the language embedding vectors to focus on different aspects, with $\gamma \geq 0$ being its weighting factor.
	
	With learned $M$ and $P=[\mu^{(1)}, \mu^{(2)}, \ldots, \mu^{(K)}]$, we compute the weights $\{\alpha_k\}_{i=1}^K$ using the unlabeled data in the target language $D_{tgt}$:
	\begin{equation}
	\label{equ:weights}
	\alpha_k=\frac{1}{\left\vert D_{tgt} \right\vert}
	\sum_{\bm{x'}\in D_{tgt}}
	\frac{\exp\left( s(\bm{x'}, \mu^{(k)})/\tau \right)}
	{\sum_{i=1}^K\exp\left( s(\bm{x'}, \mu^{(i)})/\tau \right)}
	\end{equation}
	where $\tau$ is a temperature factor to smooth the output probability distribution. 
	In our experiments, we set it as the variance of all values in $\{s(\bm{x'}, \mu^{(k)})\}, \forall \bm{x'}\in D_{tgt}, \forall k\in\{1, ..., K\}$, so that $\alpha_k$ would not be too biased to either $0$ or $1$. 
	
	\subsubsection{Teacher-Student Learning}
	\paragraph{Training: } With the combined probability distribution of entity labels from multiple teacher models, \ie, $\tilde{p}(x'_i, \Theta_T)$ in Eq.~\ref{equ:multi_teacher}, the training loss for the student model is identical to Eq.~\ref{equ:loss} and \ref{equ:train_loss}.
	
	\paragraph{Inference: } For inference on the target language, we only use the learned student model and make predictions as in the single-source scenario (Eq.~\ref{equ:infer}).

	\section{Experiments}
	\label{sec:expt}
	We conduct extensive experiments for 3 target languages (\ie, Spanish, Dutch, and German) on standard benchmark datasets, to validate the effectiveness and reasonableness of our proposed method for single- and multi-source cross lingual NER. 
	
	\subsection{Settings}
	\paragraph{Datasets}
	We use two NER benchmark datasets: CoNLL-2002 (Spanish and Dutch) \cite{tjong2002introduction}; CoNLL-2003 (English and German) \cite{tjong2003introduction}. 
	Both are annotated with 4 entity types: \texttt{PER}, \texttt{LOC}, \texttt{ORG}, and \texttt{MISC}. 
	Each language-specific dataset is split into training, development, and test sets. 
	Table~\ref{tab:dataset} reports the dataset statistics. 
	All sentences are tokenized into sequences of subwords with WordPiece~\cite{wu2016google}. Following~\citet{wu2019beto}, we also use the BIO entity labelling scheme. 
	
	In our experiments, for each \textit{source} language, an NER model is trained \textit{previously} with its corresponding labeled training set. As for the \textit{target} language, we discard the entity labels from its training set, and use it as unlabeled target-language data $D_{tgt}$. Similarly, unlabeled source-language data for learning language similarities (Eq.~\ref{equ:lang_sim}) is simulated via discarding the entity labels of each training set.
	
	\begin{table}[t]
		\centering
		\setlength{\tabcolsep}{1.5mm}
		\scalebox{0.9}{
			\begin{tabular}{c|c|c|c|c}
				\hline
				Language & Type &Train &Dev &Test  \\ \hline
				English-en &Sentence &14,987 &3,466 &3,684 \\
				(CoNLL-2003)&Entity &23,499 &5,942 &5,648 \\
				\hline
				German-de &Sentence &12,705 &3,068 &3,160 \\
				(CoNLL-2003)&Entity &11,851 &4,833 &3,673 \\
				\hline
				Spanish-es &Sentence &8,323 &1,915 &1,517 \\
				(CoNLL-2002)&Entity &18,798 &4,351 &3,558 \\
				\hline
				Dutch-nl &Sentence &15,806 &2,895 &5,195 \\
				(CoNLL-2002)&Entity &13,344 &2,616 &3,941 \\
				\hline
			\end{tabular}
		}
		\caption{Statistics of the benchmark datasets.}
		\label{tab:dataset}
	\end{table}
	
	\paragraph{Network Configurations}
	We leverage the cased multilingual $\text{BERT}_{\text{BASE}}$~\cite{wu2019beto} for both $f(\cdot)$ in Eq.~\ref{equ:h_single} and $g(\cdot)$ in Eq.~\ref{equ:similarity}, with 12 Transformer blocks, 768 hidden units, 12 self-attention head, GELU activations~\cite{dan2016bridging}, and learned positional embeddings.
	We use the final hidden vector of the first [\texttt{CLS}] token as the sentence embedding for $g(\cdot)$, and use the mean value of sentence embeddings \wrt{} the $k$-th source language to initialize $\mu^{(k)}$ in Eq.~\ref{equ:similarity}. 
	
	\paragraph{Network Training}
	We implement our proposed method based on \emph{huggingface} Transformers\footnote{https://github.com/huggingface/transformers}. 
	Following~\citet{wolf2019transformers}, we use a batch size of 32, and 3 training epochs to ensure convergence of optimization. 
	Following~\citet{wu2019beto}, we freeze the parameters of the embedding layer and the bottom three layers of $\text{BERT}_{\text{BASE}}$. 
	For the optimizers, we use AdamW~\cite{loshchilov2017fixing} with learning rate of $5e-5$ for teacher models~\cite{wolf2019transformers}, and $1e-4$ for the student model~\cite{yang2019model} to converge faster. 
	As for language similarity measuring (\ie, Eq.~\ref{equ:lang_sim}), we set $\gamma=0.01$ following~\citet{pinheiro2018unsupervised}. 
	Besides, we use a low-rank approximation for the \textit{bilinear} operator $M$, \ie, $M = U^TV$ where $U, V\in \mathbb{R}^{d\times m}$ with $d \ll m$, and we empirically set $d=64$. 
	
	\paragraph{Performance Metric}
	We use phrase level F1-score as the evaluation metric, following~\citet{tjong2002introduction}. For each experiment, we conduct 5 runs and report the average F1-score. 
	\subsection{Performance Comparison}
	
	\begin{table}[t]
		\centering
		\setlength{\tabcolsep}{1.5mm}
		\begin{tabular}{c|c|c|c}
			\hline
			&	es	&	nl	&	de \\ \hline
			\citet{tackstrom2012cross}& 59.30 & 58.40	& 40.40 \\ \hline
			\citet{tsai2016cross}& 60.55	& 61.56	& 48.12 \\ \hline
			\citet{ni2017weakly}&	65.10 &	65.40 &	58.50 \\ \hline
			\citet{mayhew2017cheap}& 65.95	& 66.50	& 59.11	\\ \hline
			\citet{xie2018neural}& 72.37	& 71.25	& 57.76	\\ \hline
			\citet{wu2019beto}$^{\dag}$&	74.50 &	79.50 &	71.10 \\ \hline
			\citet{moon2019lingua}$^{\dag}$ & 75.67 & 80.38 & 71.42\\ \hline
			\citet{wu2020enhanced}&	76.75 &	80.44 &	73.16 \\ \hline
			\hline
			Ours & \textbf{76.94} & \textbf{80.89} & \textbf{73.22} \\ \hline
		\end{tabular}
		\caption{Performance comparisons of \textbf{single-source} cross-lingual NER. $^{\dag}$ denotes the reported results \wrt. freezing the bottom three layers of $\text{BERT}_{\text{BASE}}$ as in this paper.}
		\label{tab:single_source}
	\end{table}

	\paragraph{Single-Source Cross-Lingual NER}
	Table~\ref{tab:single_source} reports the results of different single-source cross-lingual NER methods. 
	All results are obtained with English as the source language and others as target languages. 
	
	It can be seen that our proposed method outperforms the previous state-of-the-art methods. 
	Particularly, compared with the remarkable \citet{wu2019beto} and \citet{moon2019lingua}, which use nearly the same NER model as our method but is based on direct model transfer, 
	our method obtains significant and consistent improvements in F1-scores, ranging from 0.51 for Dutch to 1.80 for German. That well demonstrates the benefits of teacher-student learning over unlabeled target-language data, compared to direct model transfer. 
	Moreover, compared with the latest meta-learning based method~\cite{wu2020enhanced}, 
	our method requires much lower computational costs for both training and inference, meanwhile reaching superior performance.
	
	\begin{table}[t]
		\centering
		\begin{tabular}{c|c|c|c}
			\hline
			&	es	&	nl	&	de \\ \hline
			\citet{tackstrom2012nudging}& 61.90 & 59.90	& 36.40 \\ \hline
			\citet{rahimi2019massively}& 71.80 & 67.60 & 59.10 \\ \hline
			\citet{chen2019multi}& 73.50 & 72.40 & 56.00 \\ \hline
			\citet{moon2019lingua}$^{\dag}$ & 76.53 & \textbf{83.35} & 72.44\\ \hline
			\hline
			Ours-avg & 77.75 & 80.70 & 74.97 \\ \hline
			Ours-sim & \textbf{78.00} & 81.33 & \textbf{75.33}\\ \hline
		\end{tabular}
		\caption{Performance comparisons of \textbf{multi-source} cross-lingual NER. 
		\textbf{Ours-avg}: averaging teacher models (Eq.~\ref{equ:minimal_resource}) . \textbf{Ours-sim}: weighting teacher models with learned language similarities (Eq.~\ref{equ:weights}).
		$^{\dag}$ denotes the reported results \wrt. freezing the bottom three layers of $\text{BERT}_{\text{BASE}}$.
		}
		\label{tab:multi_source}
	\end{table}
	
	\paragraph{Multi-Source Cross-Lingual NER}
	Here we select source languages in a leave-one-out manner, \ie, all languages except the target one are regarded as source languages. For fair comparisons, we take Spanish, Dutch, and German as target languages, respectively. 

	Table~\ref{tab:multi_source} reports the results of different methods for multi-source cross-lingual NER. 
	Both our teacher-student learning methods, \ie, \textit{Ours-avg} (averaging teacher models, Eq.~\ref{equ:minimal_resource}) and \textit{Ours-sim} (weighting teacher models with learned language similarities, Eq.~\ref{equ:weights}), outperform previous state-of-the-art methods on Spanish and German by a large margin, which well demonstrates their effectiveness. We attribute the large performance gain to the teacher-student learning process to further leverage helpful information from unlabeled data in the target language. 
	Though \citet{moon2019lingua} achieves superior performance on Dutch, it is not applicable in cases where the labeled source-language data is inaccessible, and thus it still suffers from the aforementioned limitation \wrt. data availability.
	
	Moreover, compared with \textit{Ours-avg}, \textit{Ours-sim} brings consistent performance improvements. That means, if unlabeled data in source languages is available, using our proposed language similarity measuring method for weighting different teacher models can be superior to simply averaging them.

	\begin{table}[t]
		\centering
		\setlength{\tabcolsep}{1mm}
		\scalebox{0.9}{
			\begin{tabular}{c|c|c|c}
				\hline
				&	es	&	nl	&	de \\ \hline
				\hline
				\multicolumn{4}{l}{\textbf{Single-source:} }\\ \hline
				Ours & \textbf{76.94} & \textbf{80.89} & \textbf{73.22} \\ 
				HL & 76.60 (-0.34) & 80.43 (-0.46) & 72.98 (-0.24)\\  
				MT & 75.60 (-1.34) & 79.99 (-0.90) & 71.76 (-1.46)\\ \hline
				\hline
				\multicolumn{4}{l}{\textbf{Multi-source:} }\\ \hline
				Ours-avg & \textbf{77.75} & \textbf{80.70} & \textbf{74.97}\\
				HL-avg & 77.65 (-0.10) & 80.39 (-0.31) & 74.31 (-0.66) \\
				MT-avg & 77.25 (-0.50) & 80.53 (-0.17) & 74.18 (-0.79)\\ \hline
				Ours-sim & \textbf{78.00} & \textbf{81.33} & \textbf{75.33}\\ 
				HL-sim & 77.81 (-0.19) & 80.27 (-1.06) & 74.63 (-0.70) \\
				MT-sim & 77.12 (-0.88) & 80.24 (-1.09) & 74.33 (-1.00)\\ \hline
				\hline
			\end{tabular}
		}
		\caption{Ablation study of the proposed teacher-student learning method for cross-lingual NER. \textbf{HL}: Hard Label; \textbf{MT}: Direct Model Transfer; \textbf{*-avg}: averaging source-language models; \textbf{*-sim}: weighting source-language models with learned language similarities.}
		\label{tab:ablation}
	\end{table}

	\subsection{Ablation Study}
	\label{sec:ablation_study}
	

	
	\paragraph{Analyses on Teacher-Student Learning}
	To validate the reasonableness of our proposed teacher-student learning method for cross-lingual NER, we introduce the following baselines. 
	1)  \emph{Hard Label (HL)}, which rounds the probability distribution of entity labels (\ie, soft labels output by teacher models) into a one-hot labelling vector (\ie, hard labels) to guide the learning of the student model. Note that in multi-source cases, we use the combined probability distribution of multiple teacher models (Eq.~\ref{equ:multi_teacher}) to derive the hard labels. 
	To be consistent with Eq.~\ref{equ:loss}, we still adopt the MSE loss here. In fact, both MSE loss and cross-entropy loss lead to the same observation described in this subsection.
	2) \emph{Direct Model Transfer (MT)}, where NO unlabeled target-language data is available to perform teacher-student learning, and thus it degenerates into: a) directly applying the source-language model in single-source cases, or b) directly applying a weighted ensemble of source-language models in multi-source cases, with weights derived via Eq.~\ref{equ:multi_teacher} and Eq.~\ref{equ:weights}.
	
	\begin{figure*}[t]
		\centering
		\includegraphics[width=15.5cm]{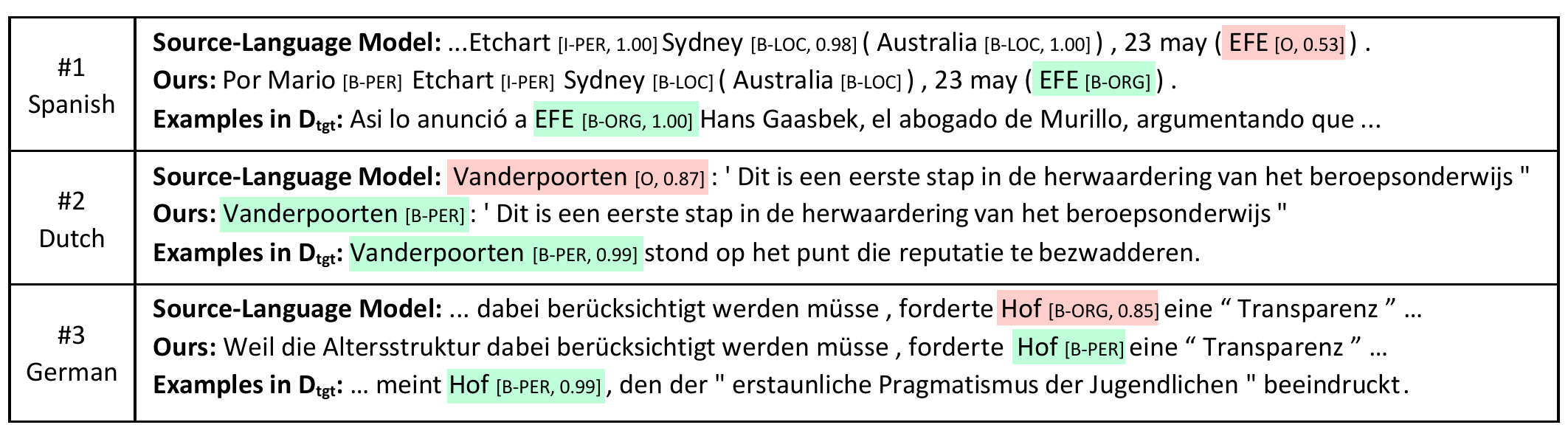}
		\caption{Case study on why teacher-student learning works. The \colorbox[rgb]{0.80, 0.98, 0.85}{GREEN} (\colorbox[rgb]{0.97, 0.82, 0.80}{RED}) highlight indicates a correct (incorrect) label. The real-valued numbers indicate the predicted probability corresponding to the entity label.}
		\label{fig:case_study}
	\end{figure*}

	Table~\ref{tab:ablation} reports the ablation study results. It can be seen that using hard labels (\ie, HL-*) would result in consistent performance drops in all cross-lingual NER settings, which validates using soft labels in our proposed teacher-student learning method can convey more information for knowledge transfer than hard labels. Moreover, we can also observe that, using direct model transfer (\ie, MT-*) would lead to even more significant performance drops in all cross-lingual NER settings (up to 1.46 F1-score). Both demonstrate that leveraging unlabeled data in the target language can be helpful, and that the proposed teacher-student learning method is capable of leveraging such information effectively for cross-lingual NER.

	\begin{table}[t]
		\centering
		\setlength{\tabcolsep}{1.5mm}
		\scalebox{0.9}{
			\begin{tabular}{c|c|c|c}
				\hline
				&	es	&	nl	&	de \\ \hline
				Ours & \textbf{78.00} & \textbf{81.33} & \textbf{75.33}\\ \hline
				$cosine$ & 77.86 (-0.14) & 79.94 (-1.39) & 75.24 (-0.09) \\ \hline
				$\ell_2$ & 77.72 (-0.28) & 79.74 (-1.59) & 75.09 (-0.24) \\ \hline
			\end{tabular}
		}
		\caption{Comparison between the proposed language similarity measuring method and the commonly used $cosine/\ell_2$ metrics for multi-source cross-lingual NER.}
		\label{tab:ablation_weights}
	\end{table}

	\paragraph{Analyses on Language Similarity Measuring}
	We further compare the proposed language similarity measuring method with other commonly used unsupervised metrics, \ie, \textit{cosine} similarity and $\ell_2$ distance. Specifically, $s(\bm{x'}, \mu^{(k)})$ in Eq.~\ref{equ:weights} is replaced by \textit{cosine} similarity or negative $\ell_2$ distance between $\bm{x'}$ and the mean value of sentence embeddings \wrt{} the $k$-th source language. 
	
	As shown in Table~\ref{tab:ablation_weights}, replacing the proposed language similarity measuring method with either \textit{cosine} / $\ell_2$ metrics leads to consistent performance drops across all target languages. This further demonstrates the benefits of our language identification based similarity measuring method. 

	\begin{figure}[t]
		\centering
		\includegraphics[width=7cm]{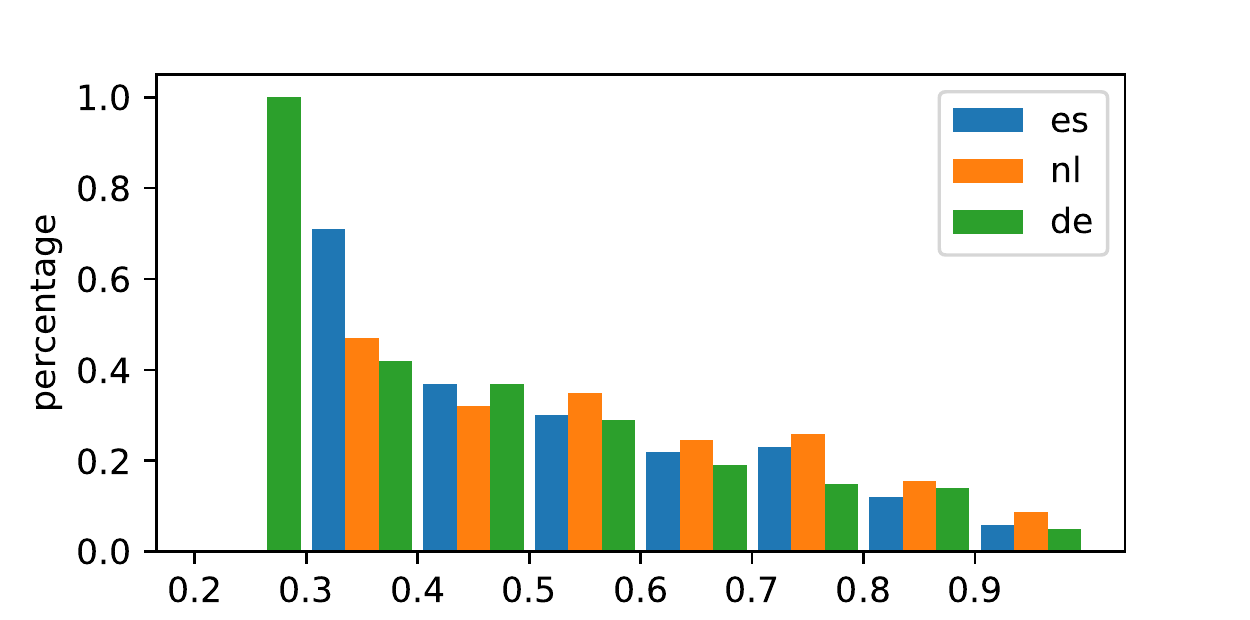}
		\caption{Percentage of corrected mispredictions, in different probability intervals.}
		\label{fig:correction}
	\end{figure}

	\subsection{Why Teacher-Student Learning Works?}
	\label{sec:case_study}
	By analyzing which failed cases of \emph{directly applying the source-language model} are corrected by the proposed teacher-student learning method, we try to bring up insights on why teacher-student learning works, in the case of single-source cross-lingual NER.
	
	Firstly, teacher-student learning can probably help to learn label preferences for some specific words in the target language. Specifically, if a word appears in the unlabeled target-language data and the teacher model consistently predicts it to be associated with an identical label with high probabilities, the student model would learn the preferred label \wrt{} that word, and predict it in cases where the sentence context may not provide enough information. Such label preference can help the predictions for tokens that are less ambiguous and generally associated with an identical entity label. As illustrated in Figure~\ref{fig:case_study}, in example \#1, the source-language (teacher) model, fails to identify ``EFE" as an \texttt{ORG} in the test sentences, while the student model (\ie, Ours) can correctly label it, because it has seen ``EFE'' labeled as \texttt{ORG} by the teacher model with high probabilities in the unlabeled target-language data $D_{tgt}$. Similar results can also be observed in example \#2 and \#3.
	
	Moreover, teacher-student learning may help to find a better classifying hyperplane for the student NER model with unlabelled target-language data. Actually, we notice that the source-language model generally makes correct label predictions with higher probabilities, and makes mispredictions with relatively lower probabilities. By calculating the proportion of its mispredictions that are corrected by our teacher-student learning method in different probability intervals, we find that our method tends to correct the low-confidence mispredictions, as illustrated in Figure~\ref{fig:correction}. We conjecture that, with the help of unlabeled target-language data, our method can probably find a  better classifying hyperplane for the student model, so that the low-confidence mispredictions, which are closer to the classifying hyperplane of the source-language model, can be clarified. 
	
	\section{Conclusion}
	In this paper, we propose a teacher-student learning method for single-/multi-source cross-lingual NER, via using source-language models as teachers to train a student model on unlabeled data in the target language. The proposed method does not rely on labelled data in the source languages and is capable of leveraging extra information in the unlabelled target-language data, which addresses the limitations of previous label-projection based and model-transfer based methods. We also propose a language similarity measuring method based on language identification, to better weight different teacher models. Extensive experiments on benchmark datasets show that our method outperforms the existing state-of-the-art approaches. 
	
	
	\bibliography{acl2020}
	\bibliographystyle{acl_natbib}
\end{document}